\documentclass[12pt]{iopart}

\usepackage{import}
\usepackage{acronym}
\usepackage{float}
\usepackage{graphicx}
\usepackage{amsmath}
\renewcommand{\thesubsection}{\arabic{subsection}}
\usepackage[normalem]{ulem}
\usepackage{xcolor}
\usepackage[switch]{lineno}

\begin{document}

\title[Continuous signal sparse encoding using analog neuromorphic variability]{Continuous signal sparse encoding using analog neuromorphic variability}

\author{Filippo Costa$^{1,2}$, Chiara De Luca$^{1,3}$}

\address{$^1$ Institute of Neuroinformatics, University of Zurich and ETH Zurich, Winterthurerstrasse 190 CH-8057 Zurich, Switzerland}
\address{$^2$ Department of Neurosurgery, University Hospital Zurich, Rämistrasse 100 Zurich, Switzerland}
\address{$^3$ Digital Society Initiative, University of Zurich, Rämistrasse 69 CH-8001 Zurich, Switzerland}
\ead{filippo.costa@usz.ch, chiara.deluca@ini.uzh.ch}
\vspace{10pt}

\begin{abstract}
Achieving fast and reliable temporal signal encoding is crucial for low-power, always-on systems. While current spike-based encoding algorithms rely on complex networks or precise timing references, simple and robust encoding models can be obtained by leveraging the intrinsic properties of analog hardware substrates.\\
We propose an encoding framework inspired by biological principles that leverages intrinsic neuronal variability to robustly encode continuous stimuli into spatio-temporal patterns, using at most one spike per neuron.\\
The encoder has low model complexity, relying on a shallow network of heterogeneous neurons. It relies on an internal time reference, allowing for continuous processing. Moreover, stimulus parameters can be linearly decoded from the spiking patterns, granting fast information retrieval.
Our approach, validated on both analog neuromorphic hardware and simulation for stimulus parameter regression and signal classification, demonstrates high robustness to noise, spike jitter, and reduced heterogeneity.
Consistently with biological observations, we observed the spontaneous emergence of patterns with stereotyped spiking order.
The proposed encoding scheme facilitates fast, robust and continuous information processing, making it well-suited for low-power, low-latency processing of temporal data on analog neuromorphic substrates.
\end{abstract}

%
\maketitle
%
\ioptwocol

\section*{Introduction}

Neuronal communication is inherently spike-based, with neurons transmitting information through sequences of action potentials (spikes) that encode sensory inputs as discrete temporal events. This spike-timing mechanism underpins the dynamic and precise nature of neural processing, enabling efficient encoding and representation of stimuli across diverse contexts~\cite{hopfield1995pattern, chong2020manipulating}. Also, neuronal responses exhibit a high degree of variability, where even neurons within the same functional column can exhibit highly variable spiking behaviors in response to identical stimuli~\cite{chelaru2008efficient, habashy2024adapting}.

Heterogeneity in neural responses is not merely a byproduct of noisy processes but a critical feature of biological systems. It has been demonstrated~\cite{chelaru2008efficient} that the high variability in intrinsic response properties of individual cells changes the structure of neuronal correlations, enhancing the information encoded in population activity and improving sensory coding~\cite{mainen1995reliability}. 
Stereotyped spiking patterns, observed in the cortex, provide a scaffold for information processing and transmission~\cite{luczak2013gating,luczak2015packet,vaz2023backbone}. These patterns represent coordinated firing across the network and can be thought of as core neural events that likely encode important information, carrying critical information about the timing and nature of sensory inputs \cite{xie2024neuronal,vaz2023backbone}. These events are thought to play a key role in compact and efficient information encoding. These dynamics highlight the potential of variability-driven mechanisms to support compact, efficient, and robust information encoding.\\

From a computational perspective, leveraging neuronal  variability and spike-based processing for encoding algorithms has significant implications for improving the efficiency and robustness of signal processing. Learning with variability has been shown to lead to more stable and robust results in simulated spiking neural networks across various tasks, especially those with rich temporal structures~\cite{perez2021neural, she2021sequence}, also allowing for the emergence of computationally specialized networks~\cite{gast2024neural, habashy2024adapting}.  
This heterogeneity also applies to the spiking encoding of sensory signals, typically continuous and temporally varying ~\cite{chelaru2008efficient, zeldenrust2021efficient}. \\

Multiple studies have explored optimal ways to exploit this spike-based processing in noisy environments to encode analog signals~\cite{gerwinn2009bayesian} and map amplitude information into time sequences~\cite{lazar2004time, lazar2009reconstruction, adam2020sampling, schrauwen2003bsa}.

Existing spike-based encoding methods demonstrated energy-efficient and low-latency information coding through precise spike timing~\cite{adam2020sampling}, providing a flexible framework to understand how neurons can effectively process temporal information in a changing environment~\cite{rabinovich2001dynamical,buonomano1995temporal}. Fast spike-based processing can be obtained with time-to-first-spike (TTFS) coding algorithms that use at most one spike per neuron but require an external time reference \cite{stanojevic2024high, goltz2021fast}. 
Reliable computation can also be obtained with recurrently-connected networks with balanced excitatory-inhibitory neurons~\cite{deneve2016efficient,boerlin2013predictive} and spatio-temporal spiking patterns~\cite{frady2019robust}.
Other approaches rely on random projections and reservoir computing~\cite{deckers2022extended} paradigms, such as Liquid State Machines (LSMs)~\cite{maass2002real}, in which high-dimensional, random transformations of the input facilitate efficient computation, leveraging recurrent neuronal dynamics and encoding temporal information in a distributed manner.\\

Building on these insights,  we propose a spike-based encoding algorithm that exploits neuronal variability to minimize network complexity and eliminates the need for external time references. Our method encodes continuous-time signals into spatio-temporal spiking patterns, with at most one spike per neuron (Fig.~\ref{fig:graphical_abstract}), using the DYNAP-SE mixed-signal neuromorphic hardware~\cite{moradi2017scalable}. \\
In contrast to previously described approaches, the proposed method does not rely on recurrent dynamics or sustained network states, as in reservoir computing models. Instead, our encoding scheme is based solely on the first spike time of each neuron. Furthermore, our method eliminates the dependency on external timing references or precise onset signals, making it inherently hardware-compatible, scalable, and computationally efficient for encoding continuous signals.
In the DYNAP-SE hardware, neurons are built using transistors in the linear domain, allowing low power consumption. However, this is also introducing neuronal variability, referred to as ``device mismatch"~\cite{zendrikov2023brain}. Although mismatch is usually thought as a source of error and inaccuracy, in our encoding scheme, the mismatch itself is providing the required variability. 
Through simulations and hardware experiments, we performed linear regression of the input stimulus parameters. We evaluated the robustness and generalizability of the proposed encoding scheme. We observed the emergence of stereotyped sequences, as also reported in cortical recordings~\cite{xie2024neuronal}, and demonstrated successful classification of the input signal type.
These findings not only underscore the relevance of variability-driven encoding for neuromorphic systems but also lay the foundation for its application in domains for which real-time processing is needed.

\section*{Results}

\begin{figure*}[!tbhp]
\centering
\includegraphics[scale=0.8]{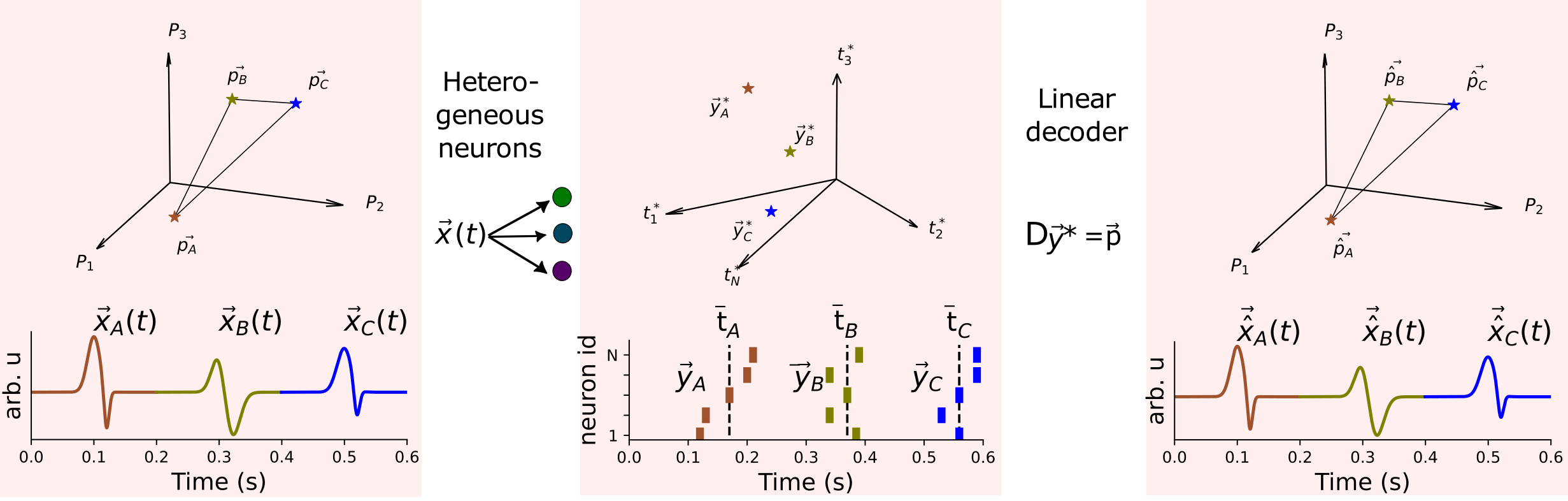}
\caption{\textbf{Continuous stimulus encoding with at most one spike per neuron}. Stimuli $\{x\}$ were sampled from the stimulus parameter space and injected into a shallow network of N neurons. Each neuron was allowed to spike at most one spike upon receiving the stimulus. Each stimulus ($\vec{x}_A(t),\vec{x}_B(t),\vec{x}_C(t)$) was encoded into an N-dimensional vector ($\vec{y}_A^*,\vec{y}_B^*,\vec{y}_C^*$) with spike times of each neuron relative to the population median (dotted lines). The N-dimensional vector was then linearly regressed to the stimulus parameters.}
\label{fig:graphical_abstract}
\end{figure*}

In this work, we propose an encoding framework mapping continuous-time stimuli into population spiking activity using a shallow network of exponential LIF neurons as summarized in Fig.~\ref{fig:graphical_abstract}. Given an input stimulus $\vec{x}(t)$, with parameter representation $\vec{p}$, and injected into the shallow network, the first spike times $t_i$ of each neuron $i$ are recorded to form the population response $\vec{y} = [t_1, t_2, \dots, t_N]$. To ensure invariance to global shifts, the median spike time $\bar{t}$ is subtracted from all spike times, producing a re-referenced encoding $\vec{y}^*$. Neurons that do not fire are assigned a value of zero. The spiking encoding is then linearly decoded to extract stimulus parameters: a linear decoder $D$ is trained to map $\vec{y}^*$ to the stimulus parameter representation $\vec{\hat{p}}$. The network optimization follows an evolutionary process, iteratively refining neuron time constants and connectivity weights to maximize decoding performance, as detailed in Methods~\ref{sec:res_algorithm}. To investigate the efficacy of our proposed encoding method, we tested its performance on both the DYNAP-SE neuromorphic hardware (see Methods~\ref{sec:method_dynapse}) and in simulated networks, testing it across different signal types (Gabor, Sinusoidal, SingleGauss, and DoubleGauss). 
Network connectivity was defined as the number of binary connections between each pre-synaptic and post-synaptic neuron. In simulations, each neuron's time constants were drawn from a normal distribution with $\sigma = 0.2$. On the DYNAP-SE hardware, this variability naturally emerged due to device mismatch~\cite{zendrikov2023brain}. To facilitate stimulus decoding, each network optimized three time constant values—membrane time constant ($\tau_{\text{mem}}$), excitatory synapse time constant ($\tau_{\text{syn}_+}$), and inhibitory synapse time constant ($\tau_{\text{syn}_-}$)—along with integer-valued synaptic weights. The individual variations in the time constants played a key role in ensuring robust and diverse population responses.

\begin{figure*}[!tbhp]
\includegraphics[width=1.\linewidth]{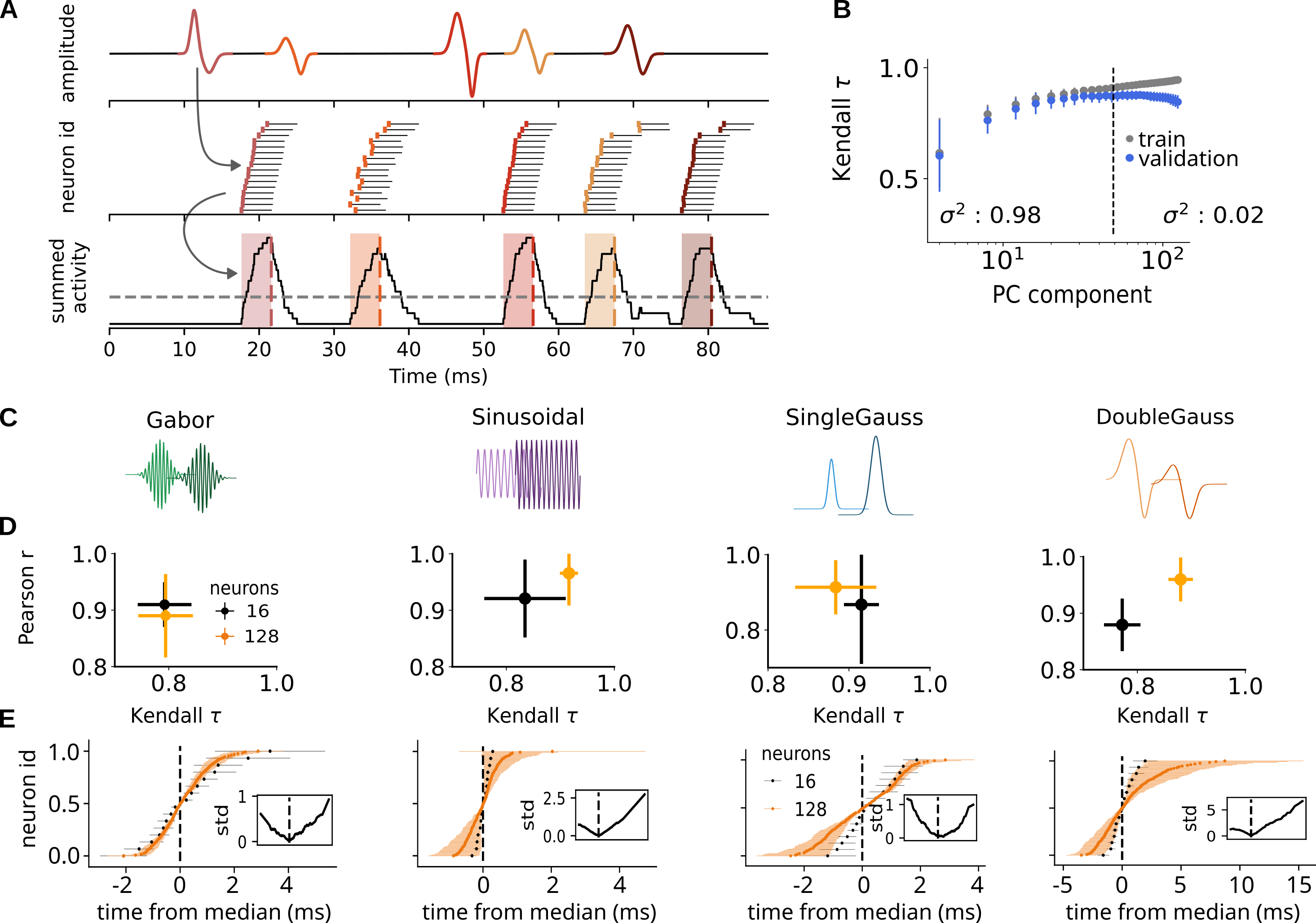}
\caption{\textbf{On-chip neural encoding performances across four synthetic signals.} 
\textbf{A)} Online signal processing evolution. A continuous ADM-encoded signal is fed into the hardware in an always-on manner (top row: six Double-Gaussian signals provided as input). The middle row shows the spiking output recorded from the DYNAP-SE chip for each neuron, sorted by first spiking time and stored in memory for a fixed time window (represented by the black horizontal line). Simultaneously, a counter tracks the rolling number of spikes occurring within the same time window, with the cumulative sum illustrated in the bottom row. When the spike count starts to decrease, spike times from the previous window (colored area) are used to compute the median value $bar{t}$ and the spiking signal representation $y^*$. \textbf{B)} Kendall-tau correlation between $\vec{p}$ and $\vec{\hat{p}}$ computed in a transformed space via PCA, as a function of the number of principal components sorted by explained variance (experiment run on populations of 128 neurons with a Double Gauss signal). While the inclusion of low-explained-variance components improves the training set score, it negatively affects generalization to the validation set, resulting in a drop in Kendall-tau correlation. The optimal number of $k$ components is selected as the value that maximizes the performance on the validation set. The averaged cumulative explained variance is shown for the first $k$ principal components ($0.98$) and for the remaining components ($0.02$), highlighting the contribution of high-explained-variance components to decoding performance. \textbf{C)} Example snippets for each synthetic signal type: Gabor, Sinusoidal, SingleGauss, and DoubleGauss. \textbf{D)} Pearson r and Kendall-tau correlation between $\vec{p}$ and $\vec{\hat{p}}$ for networks with 16 and 128 neurons. The decoding is performed using the first $k$ PC of the population response. \textbf{E)} Mean stereotyped spiking sequences for networks of 16 and 128 neurons for each input signal type, sorted in ascending order of spike time relative to the median response. Inset: standard deviation between mean sequences with the same network size as a function of the distance from the median. }
\label{fig:chip_all}
\end{figure*}

The proposed algorithm enables an always-on processing method capable of handling continuous input without requiring explicit onset information, as shown in Fig.~\ref{fig:chip_all}A. It processes signals in real time, by storing and counting neuronal spiking activity within a fixed rolling time-window. Whenever such spike count begins to decrease, the algorithm uses spike times from the preceding window to compute the median value $\bar{t}$. This value is then used to generate a relative spiking signal representation $y^*$, ensuring continuous, event-driven processing without reliance on predefined onset markers. This framework guarantees robust real-time processing of incoming signals, making it well-suited for online processing applications.
We evaluated robustness by introducing temporal jitter, spike deletions and reducing network heterogeneity, analyzing their effects on encoding accuracy.
The following sections detail the results obtained with this encoding method, including stimulus parameter regression and stimulus type classification on neuromorphic hardware and in simulations, robustness to noise and variability and the spontaneous emergence of stereotyped spiking sequences.

\subsection{Stimulus encoding with analog neurons} 
First, we implemented the proposed algorithm on neuromorphic hardware and evaluated its encoding performance on stimulus parameters regression. To ensure compatibility with the chip, stimuli were converted into spikes using an asynchronous delta modulator (ADM) before injection into the population of neurons (see Methods \ref{sec:method_adm}). We implemented the proposed stimulus encoding on the DYNAP-SE neuromorphic chip~\cite{moradi2017scalable} with 4 different synthetic signal types: ``Gabor", ``Sinusoidal",  ``SingleGauss" and ``DoubleGauss"(as shown in Fig.~\ref{fig:chip_all}C, see Methods~\ref{sec:method_signal}). Since the DYNAP-SE chip operates in real time, we opted for fast input signals with high frequencies (for ``Gabor", ``Sinusoidal" signal types) to speed up the hardware processing time. We demonstrated in simulations that the system works consistently when encoding slower signals with lower frequencies. Synaptic weights and time constants were optimized to increase the performance of the linear decoder. Decoding performance was assessed with both the Pearson's R and the Kendall-tau correlation between the stimulus parameters $\vec{p}$ and the decoded values $\vec{\hat{p}}$. Fig.~\ref{fig:chip_all} depicts the on-chip encoding algorithm performances for networks with 16 and 128 neurons, mediated over 3 different DYNAP-SE chips and 2 cores for each chip. 
We decoded stimulus parameters using the first $k$ principal components (PC) of the population response to assess the robustness of the method after network optimization and decoder training. (see Methods~\ref{sec:method_decoding}), shown in Fig.~\ref{fig:chip_all}D. 
To select the optimal first $k$ PC for the stimulus decoding, we computed the Kendall-tau correlation as a function of the number of principal components sorted by explained variance (Fig.~\ref{fig:chip_all}B). For the training set, the correlation improves monotonically as the number of PCs increases. However, for the validation set, the correlation initially improves, reaches a peak, and then decreases as more PCs are added. To select the optimal number $k$ of PCs, we choose the point where the validation performance is maximized. This analysis highlights how the PCs with low explained variance continue to increase the performance on the training set, but they become detrimental to the validation set: the additional PCs are overfitting to the training data, while failing to generalize to unseen data (validation). The selection of the number of PC to be used for testing does not rely on any prior assumptions about the signal but comes from a training and validation phases based on the measure of the optimal k value for which decoding accuracy is maximized. This approach is feasible in real-world signals as well, where we can train on available data and select the optimal number of PCs through this process.
We computed both the Pearson's R and the Kendall-tau correlation between $\vec{p}$ and $\vec{\hat{p}}$ using the first $k$ PC (Fig.~\ref{fig:chip_all}D). We obtained high correlation on all datasets (mean Pearson $r = 0.94 \pm 0.03$ s.d., mean Kendall-tau $0.88 \pm 0.05$ s.d. on the best performing network size) with small outlier percentage (mean $\% \text{ outliers} = 1.0 \pm 0.6$ s.d., see Methods~\ref{sec:method_decoding} for outlier definition). Linear and non-linear stimulus parameters were linearly decoded from the proposed encoding scheme with high accuracy. We emphasize that PC projections were computed during training, and therefore the downstream decoding was still linear during testing.
%

\subsection{On-chip stereotyped spiking sequences}
Within this framework, we investigated the spatio-temporal structure of the neural response to different stimuli. We computed the mean activity, sorted in ascending order of spike time relative to the median response, for networks of 16 and 128 neurons for all signal types. This ordering ensures a consistent representation of population activity across trials. Consistent with cortical recordings~\cite{xie2024neuronal}, we observed the emergence of stereotyped sequences, characteristic for different input types (Fig.~\ref{fig:chip_all}E). Furthermore, it was possible to observe a small variability between sequences in networks with the same size (Fig.~\ref{fig:chip_all}E-inset), increasing with the time distance from the mean. 
Also, we computed the mutation index (MI) to evaluate the similarity between spike sequences produced by the trained network (see Methods~\ref{sec:method_backbone}). This metric, recently used in neuroscience~\cite{xie2024neuronal}, is distinct from the decoding accuracy measures (Pearson r and Kendall-tau between $\vec{p}$ and $\vec{\hat{p}}$), and it does not reflect the quality of the encoder.  MI values close to 1 indicate the presence of stereotyped activity in the encoder. For all signal types and network sizes we obtained an average $\text{MI} = 0.90 \pm 0.05$. We could therefore conclude that the presented encoding scheme with analog neurons produced stereotyped spiking sequences.

\subsection{On-chip encoding robustness}
\begin{figure}
  \begin{minipage}[c]{0.4\textwidth}
    \includegraphics[width=1.2\linewidth]{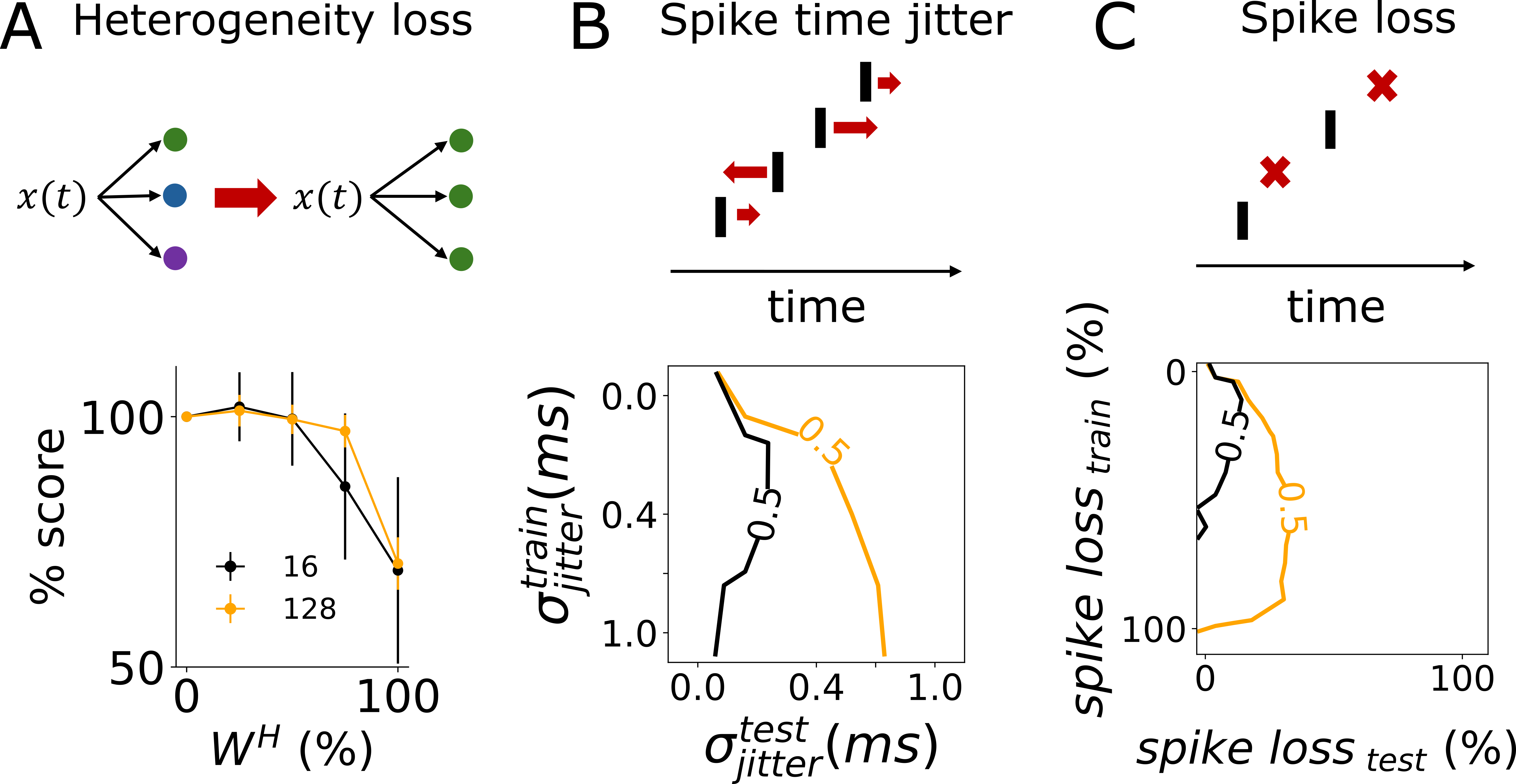}
  \end{minipage}\hfill
  \begin{minipage}[c]{0.5\textwidth}
    \caption{\textbf{Impact of neuronal variability and noise on on-chip encoding performance.} On-chip results assessing the algorithm robustness to variability reduction, temporal jitter and spike deletion on DoubleGauss signal decoding. \textbf{First row)} graphical representation of the robustness experiment. \textbf{Second row)} results from on-chip measurement for networks with 16 and 128 analog neurons (black and orange lines, respectively). For all tests, the score was computed as the percentage of the Kendall-tau correlation in the unperturbed case. \textbf{A)} Kendall-tau  correlation as a function of weight variability ($\text{w}^H(\%)$). \textbf{B)} Kendall-tau  correlation heatmap illustrating the influence of a temporal jitter applied to the output spike trains during the training and testing phases. 
\textbf{C)} Kendall-tau correlation score heatmap illustrating the influence of spike deletion in the output spike trains applied during the training and testing phases.} \label{fig:robustness}
  \end{minipage}
\end{figure}

Neuronal variability is the core of the proposed algorithm. To quantitatively evaluate the robustness to various amounts of variability and noise of such algorithm implemented on-chip, we proceeded as follows. First, to investigate how variability relates to encoding performance, we trained the linear decoder over the activity of progressively more homogeneous networks implemented on-chip,  increasing the percentage of weight sharing between neurons ($w^H$). As expected, decreasing the network variability leads to worse encoding performances (Fig.~\ref{fig:robustness}A). Comparing networks with 16 and 128 neurons, it appears that both networks had similar behavior for the same absolute amount of variability. Considering relative variability reduction, larger networks were more robust. We show that heterogeneity in the network has a clear benefit on encoding performance, as indicated by the degradation of performance when variability is reduced. Moreover, we demonstrate that even when enforcing full weight homogeneity on-chip, the hardware itself introduces intrinsic variability, as performance does not drop to zero but preserves more than 60\% of the performance of the fully heterogeneous network.
Furthermore, in Fig.~\ref{fig:robustness}B we evaluated the robustness of the encoded representation against random time jitters  in the output spike trains, taking inspiration from the noisy communication within the brain. This has been tested by perturbing the spiking activity with gaussian noise sampled from $\mathcal{N}(0,\sigma)$ with increasing $\sigma$ (from 0 to 1 ms). If noise was introduced already within the training set, test decoding was more robust to noisy activity than if noise was only included in test sets. We also tested the network robustness to noise introduced in the form of spike deletion in the output spike train (Fig~\ref{fig:robustness}C). In this test, the robustness increased with the network size.

\subsection{Size and time constant analysis on simulated networks} 
In order to have full control on neuronal dynamics and variability, towards a complete understanding of the algorithm behavior and limitations, we have also evaluated such an algorithm within a simulated framework. 

\begin{figure*}[!tbhp]
\centering
\includegraphics[width=1.\linewidth]{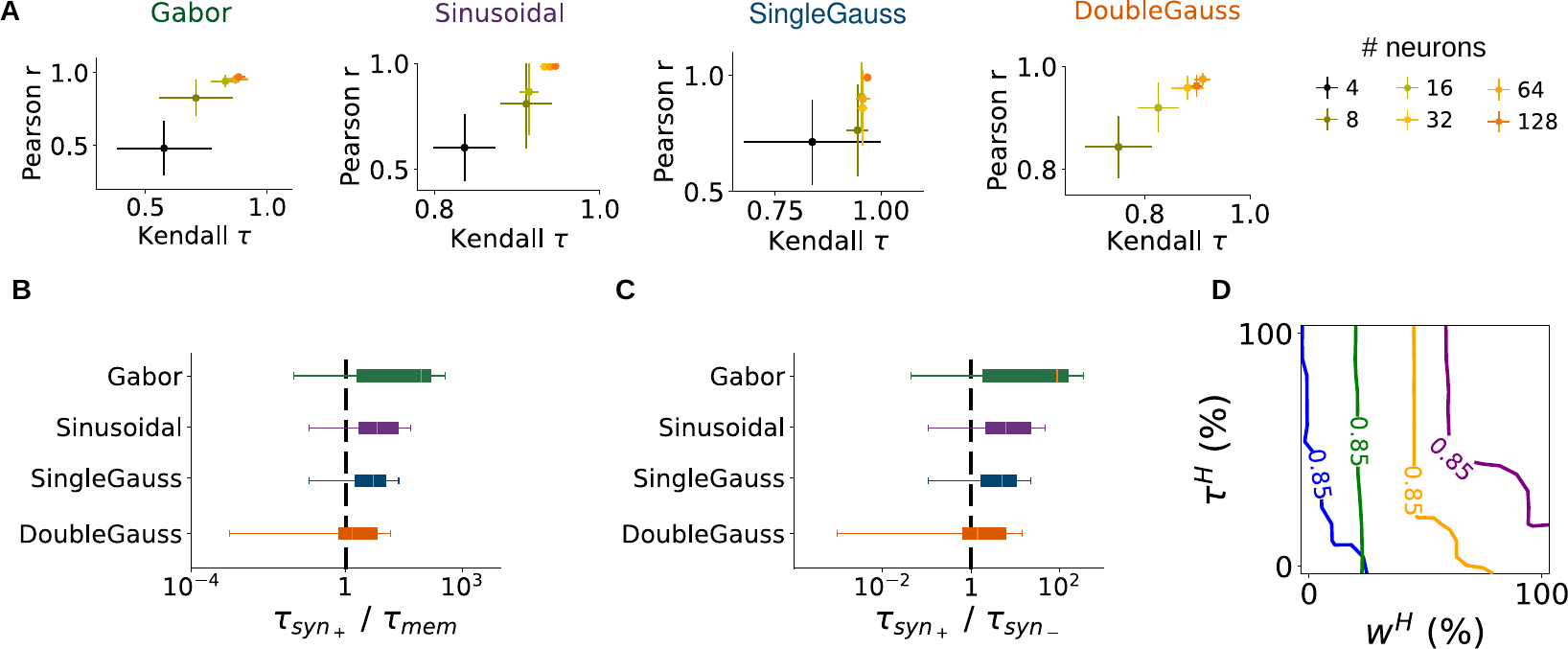}
\caption{\textbf{Enconding algorithm characterization on simulated networks.}
Single-spike encoding algorithm leveraging the variability of a simulated shallow network of exp-LIF neurons tested on the same input signal types used for on-chip testing. \textbf{A)} Kendall-tau and Pearson r correlation between original and decoded stimulus parameters for networks with an increasing number of neurons. Stimulus decoding with the first $k$ principal components (PC) consistently showed improved correlation as the network size increased.\textbf{B)} The parameter ratio between the excitatory synapses time constant ($\tau_{syn_{+}}$) and the membrane time constant ($\tau_{mem}$) averaged over multiple runs across multiple network sizes for the four signal types. The mean ratio consistently remains above 1 for all signal types. \textbf{C)} The parameter ratio between the excitatory synapses time constant ($\tau_{syn_{+}}$) and the inhibitory synapses time constant ($\tau_{syn_{-}}$) averaged over multiple runs across multiple network sizes for the four signal types. The mean ratio consistently remains above 1 for all signal types. \textbf{D) Contour map showing the regions where the} Kendall-tau correlation heatmap for each signal type remained above 0.85, as a function of weight variability ($\text{w}^{H}(\%)$) and time constant variability ($\tau^{H}(\%)$). Larger contour areas indicate smaller sensitivity to variability, with weight variability generally having a stronger impact than time constant variability.}
\label{fig:simulation_results}
\end{figure*}

Specifically, we implemented simulations of networks of exp-LIF neurons with inhibitory and excitatory synapses with heterogeneous time constants and weights (see Methods~\ref{sec:method_neuronmodel}). In a simulated environment we could monitor time constant values for different signal types and network sizes. As shown in Fig.~\ref{fig:simulation_results}A, optimizing simulated networks with a wide array of network sizes, we observed an increase in Kendall-tau correlation with network size when decoding using the first $k$ PC of the population response. Decoding with PC, we obtained a mean Kendall correlation $0.93 \pm 0.03$ s.d. on the best performing network size. Analogously, the Pearson correlation between $\vec{p}$ and $\vec{\hat{p}}$ increased with the network size. We obtained a mean Pearson correlation $ r = 0.92 \pm 0.03$ s.d. with $\%$ outliers $ = 0.2 \pm 0.2$ s.d. on the best performing network size. We then evaluated the ratio between time constant values for different signal types (Fig.~\ref{fig:simulation_results}B and C). These observables showed that, on average, for each signal type the optimization procedure converged to $\tau_{\text{syn}_+} \geq \tau_{\text{syn}_-} \geq \tau_{\text{mem}}$.
Being able to control both weight and time constant variability, we expanded the variability robustness results shown in Fig.~\ref{fig:robustness}A and tested which kind of variability drives encoding performance the most (Fig.~\ref{fig:simulation_results}D). For all signal types, weights had a stronger effect on performance than time constants. However, optimizing both types of variability led to the highest encoding performances.
To test the presence of stable spiking patterns in simulated networks, we computed their cosine similarity for different network sizes for each signal type. Considering all signal types, we obtained a mean $0.8 \pm 0.1$ cosine similarity: spiking patterns were therefore stable for different network sizes.

\subsection{Stereotyped spiking sequences are signal-type specific}
\begin{figure}
  \begin{minipage}[c]{0.4\textwidth}
  \includegraphics[scale=0.25]{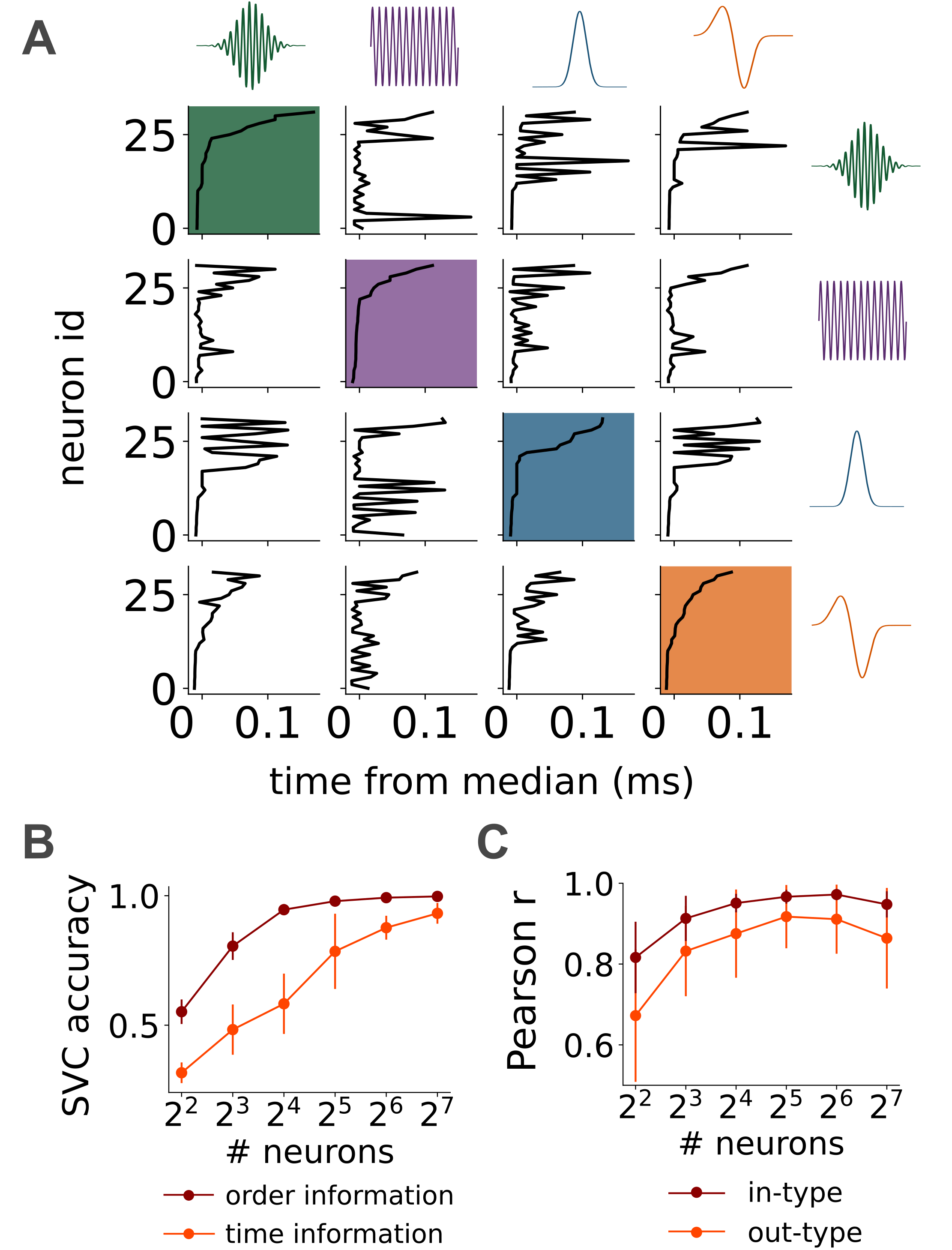}
  \end{minipage}\hfill
  \begin{minipage}[c]{0.5\textwidth}
    \caption{\textbf{Stimulus-specific spiking patterns in one simulated network.} \textbf{A)} We injected stimuli from all 4 signal types into one simulated trained network and obtained the mean spiking order for each stimulus type. Each row represents the mean spiking activity for stimulus type $i$, sorted according to the mean spiking order of stimulus type $j$. \textbf{B)} Classification of stimulus type from the encoding of one network using time and order information. For both information types we observed an increase in classification accuracy when increasing the network size. Order information produced a classification accuracy higher than the one obtained with time information. \textbf{C)} Linear decoding of stimulus parameters for all stimulus types in one network. Pearson $r$ was higher for the stimulus type the network has been trained on (in-type) compared to the mean of the other stimulus types (out-type). On average, Pearson $r$ increased with the network size till a plateau was reached.}
    \label{fig:backbone}
  \end{minipage}
\end{figure}

When presenting visual stimuli to humans, cortical spiking sequences appear to be stimulus-type specific \cite{xie2024neuronal}. We tested if networks trained on one stimulus type were able to produce different patterns for other types of stimuli. Injecting all 4 stimulus types into simulated trained networks, we observed the presence of 4 different stereotyped sequences. (Fig.~\ref{fig:backbone}A). 

Motivated by this qualitative result, we then tested if stimulus type could be linearly classified at the single trial level from the proposed encoding. Using a linear Support Vector Classifier (SVC), we classified the input signal type using both time information and order information. We define as classification with time information a linear SVC that considers the $N$-dimensional spiking encoding $\vec{y^\star}$, with $N$ the number of neurons. We define as classification with order information a linear SVC that considers a binary vector of dimension $N(N-1)/2$. This vector contains the spiking order information between all possible pairs of neurons (see Methods \ref{sec:method_classification}) On average, we observed an increase in the decoding accuracy when increasing the network size for both classifications. Using order information, we consistently observed higher classification accuracy (Fig.~\ref{fig:backbone}B).

We then assessed the linear decoding accuracy of stimulus parameters for all stimulus types in one network (Fig.~\ref{fig:backbone}C). For all networks, the Pearson correlation was higher for the stimulus type the network has been trained on (in-type) compared to all the other types (out-type). On average, increasing the network size we achieved higher decoding accuracy (Fig.\ref{fig:simulation_results}).

\subsection{Shift-invariant signal classification on-chip}
\begin{figure}
    \centering
    \includegraphics[width=1.\linewidth]{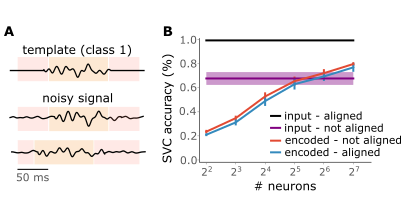}
    \caption{\textbf{Classification accuracy of a linear SVC on continuous and encoded data with and without temporal alignment.} \textbf{A)} Example of a class template and two aligned and not aligned corresponding noisy signals used for classification. The signal consists of a $100~ms$ template combined with additional filtered Gaussian noise. \textbf{B)} Classification accuracy as a function of the number of neurons used in the spiking encoder. The SVC over the input signal achieves perfect accuracy on the aligned dataset but drops below 80\% when temporal shifts are introduced (not aligned). In contrast, the spiking encoder exhibits shift-invariant properties, with accuracy improving as the number of neurons increases. This highlights the robustness of the spiking encoding approach in scenarios with temporal variability. Accuracy computed as the mean over 200 train-test splits with 400 train samples and 100 test samples.}
    \label{fig:classification}
\end{figure}

To evaluate the performance of a spiking encoder that only uses order information to classify signal types, we tested the classification accuracy on datasets composed of 6,8 and 10 classes of Gaussian noise signals filtered in the 10-100 Hz frequency band. We compared the classification performance of a SVC that uses the continuous non-encoded input and the encoded signals under both temporally  aligned and not aligned conditions (see Methods~\ref{sec:method_signal}). As shown in Fig.~\ref{fig:classification}, the SVC on the input signal achieved perfect classification accuracy on the aligned dataset but exhibited a significant drop in accuracy (below 80\%) when applied to the not aligned dataset. This suggests that the linear SVC that uses the continuous non-encoded signal relies on precise temporal alignment for classification. In contrast, our spiking encoder, implemented on the DYNAP-SE neuromorphic hardware, demonstrated shift-invariant properties. While SVC accuracy over the encoded signal did not reach perfect classification, it steadily improved as the number of neurons increased and it was not affected by the time shifts, indicating the benefits of this encoding mechanism. This shift-invariance arises from the online nature of the encoding process, which does not depend on specific temporal onsets. These results highlight the potential of this spike-based encoding methods in scenarios where robustness to temporal shifts is crucial, such as real-time signal processing applications. Such results are proven stable for classification tasks with 6,8 and 10 classes, with SVC classification accuracy over the encoded signal of $0.80 \pm 0.02$, $0.74 \pm 0.03$, $0.70 \pm 0.06$ for 128 neurons respectively (mean $\pm$ s.d.).

\section*{Discussion}
This work introduced an algorithm for continuous-time signal encoding that leverages the variability of exp-LIF neuron populations to produce robust spatio-temporal patterns with at most one spike per neuron. While our method transforms the input into a high-dimensional representation before performing linear regression—similar to reservoir computing approaches~\cite{maass2002real}—it differs fundamentally from those methods. First, conventional methods typically accumulate spiking activity over time to create a dynamic firing rate representation, whereas our approach encodes information solely through the timing of the first spike. This results in a sparse, event-driven representation that reduces computational complexity while preserving critical stimulus information. Second, reservoir computing approaches often require recurrent connectivity, increasing hardware implementation demands. In contrast, our encoding method eliminates the need for recurrent processing, ensuring compatibility with low-power, mixed-signal neuromorphic hardware such as DYNAP-SE. Finally, whereas conventional approaches scale with both the number of neurons and the duration of the signal, our method scales with the number of processing units, making it particularly well-suited for real-time, always-on processing of continuous signals.
Furthermore, our approach builds on the principles of neuronal variability and spike-based encoding~\cite{deneve2016efficient,zeldenrust2021efficient}, while minimizing network complexity by exploiting the intrinsic variability of the analog neuromorphic hardware. These distinctions underscore the efficiency and applicability of our method for neuromorphic computing. Using neuronal variability and precise spike timing, we offer a hardware-friendly alternative to traditional reservoir computing approaches while maintaining robust encoding capabilities.
By aligning with the observed features of some cortical circuits~\cite{luczak2015packet, vaz2023backbone, xie2024neuronal}, where stimulus information is encoded into precise spatio-temporal spiking patterns, this method bridges theoretical neuroscience and practical neuromorphic implementations.
Our results demonstrate that the spiking encoder effectively classifies noisy, general signals by leveraging activation order sequences, achieving robustness even in the presence of significant background noise. Furthermore, our encoding method remains invariant to temporal shifts, highlighting its potential for real-world applications where precise timing information may be unreliable or inconsistent.

The proposed method demonstrates robust performance, supporting the linear decoding of multiple parameters, both linear and nonlinear. The compatibility with the DYNAP-SE mixed-signal neuromorphic platform highlights its adaptability to challenging neuromorphic hardware. Furthermore, the insights gained from synthetic simulations extend to hybrid analog-digital systems and emerging unconventional computing frameworks. Future work will explore the application of this algorithm to other spiking processors, which may provide additional flexibility and scalability.

The algorithm employs the median spike time as an internal clock, avoiding dependence on external time references or stimulus onset markers~\cite{goltz2021fast, stanojevic2024high}. This population-driven timing aligns with biological evidence of reliable temporal coding through population activity~\cite{chase2007first} and gives always-on capabilities to the encoder. The compact, spike-based representations and the linear downstream decoding not only minimize computational overhead but also enable fast information retrieval. 

While this study validates the algorithm through synthetic simulations and implementation on DYNAP-SE hardware, it has not been extensively tested with real-world data. However, our findings establish a strong foundation for its applicability in diverse signal processing scenarios, demonstrating robustness across a broad frequency range and noisy conditions. Future research will focus on applying the method to practical scenarios as biomedical signal processing, where additional calibration in terms of the number of neurons and the degree of heterogeneity may be required to optimize performance for specific signal types. Another limitation is the current single-spike-per-neuron constraint, which, while computationally efficient, may limit the richness of encoded representations.
Future research may examine the impact of relaxing this constraint to allow for more flexible encoding schemes, tailored to specific tasks.

In summary, this study presents a robust, brain-inspired method for encoding continuous stimuli into compact spike-based representations. By leveraging neuronal variability and precise spike timing, this low-complexity approach offers low-latency, always-on capabilities for neuromorphic signal processing and bridges the gap between biological principles, computational models, and technological applications.

\section*{Methods}
\setcounter{subsection}{0}
\subsection{The algorithm}
\label{sec:res_algorithm}
Each continuous-time stimulus input to the network can be represented in two distinct ways: as a time-series function $\vec{x}(t)$, which describes its temporal evolution, and as a parameter-based representation $\vec{p}$, which characterizes its underlying structure. Given a input signal type defined by a function $f(\vec{p}, t)$ (see Methods~\ref{sec:method_signal}) then the two representations are connected as $\vec{x}(t) = f(\vec{p}, t)$. The proposed algorithm used a shallow network of exponential LIF neurons trained to extract the parameter representation $\vec{p}$ from the temporal signal $\vec{x}(t)$ (Fig.~\ref{fig:graphical_abstract}).
When a stimulus $\vec{x}(t)$ is presented to the network, each neuron $i$ responds with a first spike at time $t_i$, forming the population response $\vec{y} = [t_1, \dots, t_N]$. Since absolute spike times can vary due to global shifts, we compute the median spike-time $\bar{t}$ as a reference. The response is then re-referenced by subtracting this median from each spike-time, yielding the encoded pattern $\vec{y}^* = [t_1^*, \dots, t_N^*]$ with $t_i^* = t_i - \bar{t}$. This transformation ensures that the encoding captures the relative timing between spikes rather than absolute spike times, making it invariant to global shifts. If a neuron did not fire after stimulus injection, its corresponding value was set to 0 after the re-referencing.
The network was therefore encoding $\vec{x}(t)$ into the temporal population coding $y^*$. A linear decoder $D$ was then trained to map $\vec{y}^*$ to the parameter representation $\vec{\hat{p}}$ yielding an estimate $\vec{\hat{p}} = D y^*$. 
During training (see Methods~\ref{sec:method_optimization}), the decoding performance was evaluated using the Kendall-tau correlation between the ground-truth parameters $\vec{p}$ and the decoded representation $\vec{\hat{p}}$.\\
Such algorithm was implemented on both the DYNAP-SE neuromorphic hardware and on simulated networks, under three main conditions:
\begin{itemize}
    \item Before injecting the stimulus into the network, the continuous signal was converted into a 2-dimensional spike stream with an Asynchronous Delta Modulator (ADM, see Methods~\ref{sec:method_adm} for details).
    \item Weights were defined as the number of connections between each pre-synaptic neuron to each post-synaptic one.
    \item Each neuron was characterized by an intrinsic time constant variability. Therefore, each neuron $i$ had its own time constant value defied as $\tau_i = \tau + \eta^{\tau}_i * \tau$. On simulated networks $\eta^{\tau}_i$ was drawn from $\mathcal{N}(0,\sigma)$ with $\sigma = 0.2$. On DYNAP-SE the variability was provided by device mismatch, and it was shown to be comparable to the one implemented in simulations~\cite{zendrikov2023brain}. $\tau$ was shared between neurons.
\end{itemize}
The proposed encoding algorithm was implemented by a shallow network of $N$ neurons with weight and time constants variability. To decode stimulus parameters, each network could optimize 3 time constants values (membrane time constant $\tau_{\text{mem}}$, excitatory synapses $\tau_{\text{syn}_+}$ and inhibitory synapses $\tau_{\text{syn}_-}$ time constants) and $4N$ integer weights, for a total of $3 + 4N$ parameters. 
As described in Methods~\ref{sec:method_optimization}, the network optimization proceeded iteratively, using a simple evolutionary algorithm. Time constants were randomly sampled from a spherical sampling volume ~\cite{costa2024robust}. The decoding score for each network configuration was computed and the best combination of time constant values was selected as the center of the sampling volume for the next iteration of the algorithm. Network weights were progressively modified and tested. If the new network connectivity increased the score, the improved configuration was retained for further evolution.\\
\subsection{DYNAP-SE} 
\label{sec:method_dynapse}
DYNAP-SE is a multi-core asynchronous mixed signal neuromorphic processor. Each of the 4 cores comprises 256 adaptive exponential Leaky Integrate \& Fire (LIF) silicon neurons with two excitatory and two inhibitory analog synapses.
In the limit of high input current and shutting off the adaptation, the neural dynamics of the DYNAP-SE LIF can be expressed as:

$$\tau \frac{d}{dt} I_{\text{mem}} + I_{\text{mem}} \approx \frac{I_{\text{in}} I_{\text{gain}}}{I_{\tau}} + \frac{I_a I_{\text{mem}}}{I_{\tau}} $$
where $I_\text{mem}$ is the membrane potential, $\tau$ is the neuron time constant, and $\frac{I_a I_{\text{mem}}}{I_{\tau}}$ models the positive feedback block. Each neuron has a 64 connections fan-in and 1024 connections fan-out.
Neurons in each DYNAP-SE core share bias settings and, therefore, time constant values are shared.

\subsection{Neuron model on simulated network}
\label{sec:method_neuronmodel}
We used the exponential leaky integrate-and-fire neuron model for the simulated networks. The neuron has a membrane time constant defined as $\tau_{\text{mem}}$, firing when the membrane potential $V(t)$ reaches the threshold $\theta$. The neuron dynamics is defined as:

\begin{align*} 
    \frac{dI_{+}(t)}{dt} &= - \frac{I_{+}(t)}{\tau_{\text{syn}+}}  + w_{+}^{UP} \sum_i \delta(t-t^{UP}_{i})\ \\&  \mspace{20mu} + w_{+}^{DN} \sum_i \delta(t-t^{DN}_{i})\\
    \frac{dI_{-}(t)}{dt} &= -\frac{I_{-}(t)}{\tau_{\text{syn}-}}  + w_{-}^{UP} \sum_i \delta(t-t^{UP}_{i})\ \\ &\mspace{20mu}  + w_{-}^{DN} \sum_i \delta(t-t^{DN}_{i})\\
    I_{\text{syn}}(t)    &= -\ I_{-}(t) +I_{+}(t) \\
    \frac{dV(t)}{dt}    &= -\frac{V(t)}{\tau_{\text{mem}}} + I_{syn}(t).\\    
\end{align*}
Each neuron $i$ can fire at most one spike when its voltage threshold is reached. To address chip compatibility, synaptic weights $w$ can assume only integer values. The neurons have heterogeneous properties. In a population of $N$ neurons, only one value can be assigned for each time constant ($\tau_{\text{mem}}, \tau_{\text{syn}+}, \tau_{\text{syn}-}$). Each simulated neuron $i$ has an intrinsic variability given by a multiplicative noise term. For a given neural time constant $\tau$, neuron $i$ presents the following parameter value: $\tau_i = \tau + \eta_i^{\tau} * \tau$ where $\eta_i^{\tau} \sim \mathcal{N}(0,\sigma)$ with $\sigma = 0.2$. The results presented an average across $10$ different neural population instances.

All simulated neurons are provided with  two kinds of synapses: one excitatory with synaptic time constant $\tau_{\text{syn}+}$ and one inhibitory with synaptic time constant $\tau_{\text{syn}-}$. On DYNAP-SE, when a large input current is provided, neurons follow the same dynamics as in the simulation~\cite{chicca2014neuromorphic} but with a current-based positive feedback that drives spike generation.

The simulation framework we developed allows for customizable neuron models, network configurations, and variability conditions, making it applicable to a broad range of neuromorphic systems. By simulating different levels of heterogeneity and noise, it provides a robust testing environment for evaluating encoding algorithms beyond the DYNAP-SE hardware.

\subsection{Datasets} 
\label{sec:method_signal}
On DYNAP-SE and on simulated networks, each stimulus is represented as a time-series function $\vec{x}(t)$ generated by a parameterized function $f(\vec{p}, t)$, where $\vec{p}$ denotes the set of parameters that define the stimulus. In this work, we considered 4 signal types:
\begin{itemize}
    \item Sinusoidal stimulus:
    \begin{equation*}
            \vec{x}(t) = f_{\text{A}}(\vec{p}, t) = p_1 \text{sin}( 2\pi p_2 t )
    \end{equation*} 
    with $p_1 \in [5,10), \quad p_2 \in [500,1500) \text{Hz}$ on DYNAP-SE and $p_1 \in [1,6), \quad p_2 \in [10,100) \text{Hz}$ on simulation.
    
    \item Gabor stimulus:
    \begin{equation*}
            \vec{x}(t) = f_{\text{B}}(\vec{p}, t) = 3e^{-\frac{t^2}{2p_1^2}}\text{sin}(2\pi p_2 t)
    \end{equation*} 
        
        with $p_1 \in [2,3)\ \text{ms}, \quad p_2 \in [500,1500)\ \text{Hz}$ on DYNAP-SE and $p_1 \in [20,40)\ \text{ms}, \quad p_2 \in [10,100)\  \text{Hz}$ on simulation.
    \item SingleGauss stimulus:
    \begin{equation*}
            \vec{x}(t) = f_{\text{C}}(\vec{p}, t) = p_1 e^{-\frac{t^2}{2p_2^2}}
    \end{equation*}   
        with $p_1 \in [1,6), \quad p_2 \in [0.2,1.3)\ \text{ms}$ on DYNAP-SE and $p_1 \in [1,6), \quad p_2 \in [10,30)\ \text{ms}$ on simulation.
    \item DoubleGauss stimulus:
    \begin{equation*}
            \vec{x}(t) = f_{\text{D}}(\vec{p}, t) = p_1e^{-\frac{t^2}{2p_2^2}} + p_3e^{-\frac{(t-0.02)^2}{2p_4^2}}
    \end{equation*} 
    with $p_1, p_3 \in [1,3), \quad p_2,p_4 \in [0.6,1)\ \text{ms}$ on DYNAP-SE and $p_1,p_3 \in [1,3), \quad p_2,p_4 \in [4,10)\ \text{ms}$ on simulation.
\end{itemize}
On DYNAP-SE, each stimulus injection lasts for $10$ ms with a sampling frequency $f_s = 5e4$. On simulation, each stimulus injection lasts for $200$ ms with a sampling frequency $f_s = 5e3$.

Sampling frequency on DYNAP-SE is chosen to allow fast processing. Parameters are chosen to guarantee optimal ADM conversion with the predefined sampling frequency.

Also, to evaluate the performance of our spiking encoder on a signal classification task with general signals, we constructed a dataset based on filtered Gaussian noise signals designed to test classification robustness under temporal shifts. First, we generated 6 different "template" signals, each corresponding to a class. These templates were created by extracting 100 ms of Gaussian noise and filtering it within the [10, 100] Hz frequency band with an amplitude of 1. To generate training and testing examples for each class, we added additional noise components to the templates. Specifically, each class example consisted of the template signal plus a second independent 100 ms extraction of Gaussian noise, filtered in the same [10, 100] Hz band but with a reduced amplitude of 0.5. Additionally, we introduced a broader noise background by overlaying 200 ms of Gaussian noise, filtered within the same frequency band but with a lower amplitude of 0.2. This ensured variability while preserving class-distinctive features. To test temporal robustness, we also created a shifted version of the dataset in which each example was randomly shifted in time within a range of [-20, 20] ms. This allowed us to compare classification performance under both aligned and not aligned conditions, assessing the shift-invariance of the encoding methods.

\subsection{ADM converter} 
\label{sec:method_adm}
Stimuli are converted into spikes using an asynchronous delta modulator (ADM) before injection into a population of neurons.  
For each stimulus $x(t)$, we consider its reference value $x(0)$ and two thresholds at $x(0) + \delta$ (UP threshold) and $x(0) - \delta$ (DN threshold). If the signal at time $t^{*}$ crosses the UP (DN) threshold, a UP (DN) event is generated at time $t^{*}$, and the UP/DN thresholds are updated as $x(t^{*}) + \delta$ and $x(t^{*}) - \delta$. This process converts each stimulus into a stream of UP and DN events. We can then obtain the ADM-reconstructed version of the stimulus $\Tilde{x}(t)$ from the stream of UP/DN events. From $\Tilde{x}(0) = 0$, we update the stimulus amplitude every time a UP/DN event occurs. If a UP event occurs at time $t^{*}$, then $\Tilde{x}(t^{*}) = \Tilde{x}(t^{*}-1) + \delta$ ($- \delta$ for a DN event).
The ADM threshold is not chosen based on the final accuracy of the encoding signal but rather by optimizing the reconstruction of the input signal. The optimal ADM threshold value is determined by maximizing the Pearson correlation between Euclidean distances of stimuli computed from the continuous dataset and their reconstructed version. Additionally, it's important to note that the ADM conversion is primarily used to interface the analog input signal with the DYNAP-SE hardware.

\subsection{Regression evaluation metrics}
\label{sec:methods_metrics}
To quantify the accuracy of the stimulus parameter regression, we used two complementary metrics: Kendall’s Tau and the Pearson’s r correlation coefficient.\\

Kendall’s Tau is a non-parametric measure that quantifies the rank correlation between the set of stimulus parameters $\vec{p}$ and the set of decoded values $\vec{\hat{p}}$. It is defined as:
\begin{equation}
\tau = \frac{C - D}{\frac{1}{2} N (N - 1)}
\end{equation}

where:
\begin{itemize}
    \item $C$ is the number of concordant pairs,
    \item $D$ is the number of discordant pairs,
    \item $N$ is the total number of observations.
\end{itemize}

This metric is robust to outliers, and  was used to optimize the shallow network of heterogeneous neurons.

Pearson’s r correlation coefficient quantifies the linear relationship between the set of stimulus parameters $\vec{p}$ and the set of decoded values $\vec{\hat{p}}$:

\begin{equation}
r = \frac{\sum_{i} (p_i - \bar{p}) (\hat{p}_i - \bar{\hat{p}})}{\sqrt{\sum_{i} (p_i - \bar{p})^2} \sqrt{\sum_{i} (\hat{p}_i - \bar{\hat{p}})^2}}
\end{equation}

where:
\begin{itemize}
    \item $p_i$ and $\hat{p_i}$ are the values of parameter $p_i$ and its reconstruction $\hat{p}_{i}$,
    \item $\bar{p}$ and $\bar{\hat{p}}$ are their respective means.
\end{itemize}

This metric was used to quantify the deocding accuracy of stimulus parameters after the network optimization stage.\\

\subsection{Encoding algorithm}
\label{sec:method_packet_based}
Stimulus $\vec{x}(t)$  has a parameter representation $\vec{p}$ $ = (p_1 \dots p_K)$.  When the ADM-converted stimulus $\vec{x}(t)$ is injected into a population of $N$ neurons, a spiking output $\vec{y}(t) = [t_{1}, \dots t_{N}]$  is produced, where $t_i$ is equal to the firing time of neuron $i$. If neuron $j$ does not fire, then $t_j = 0$. To make the representation invariant to global timing shifts, the spike times are then re-referenced to the median $\bar{t}$ of active neurons' firing times. The stimulus is then mapped to an N-dimensional vector $\vec{y}^*$ defined as:

\begin{equation*}
    t^*_i =
    \begin{cases}
        t_i - \bar{t} & \text{if } t_i \neq 0,\\
        0       & \text{if } t_i = 0.
    \end{cases}
\end{equation*}

\subsection{Linear decoding}
\label{sec:method_decoding}
A linear regression is performed from the encoded signal $\vec{y^*}$ to the stimulus parameters $\vec{p} = (p_1 \dots p_K)$. The estimated parameters are denoted as $\hat{p} = (\hat{p}_1, \dots, \hat{p}_K)$. To assess the performance of the linear decoder, we compute the Kendall-tau correlation between the true parameters $\vec{p}$ and the decoded parameters $\vec{\hat{p}}$. We also compute the Pearson correlation between  $\vec{p}$ and $\vec{\hat{p}}$ after removing outliers. Outliers are defined as values of $\hat{p}$ that are either greater than twice the largest  $\vec{p}$ or smaller than half the smallest $\vec{p}$.

When decoding using principal components, we performed the following steps: we first performed PCA on the population response of the training set; we then assessed the decoding accuracy on a validation set, progressively increasing the number of principal components. The first $k$ principal components for which we obtained the highest validation accuracy were then used for testing. Since we performed PCA only during training, and we kept the linear projection fixed during testing, the final downstream decoding is still linear. 

\subsection{Optimization protocol}
\label{sec:method_optimization}
Network optimization follows two main steps:
\begin{itemize}
    \item Neural time constants $\{\tau_{mem}, \tau_{\text{syn}+}, \tau_{\text{syn}-}\}$ are sampled from a uniform distribution with radius $r$. The Kendall-tau correlation of each network configuration is evaluated, and the best-performing set of time constants is set as the center of the new sampling space.
    \item Weights are integer values that can be changed as follows: First, a neuron index $i$ is randomly extracted. This neuron can then change its excitatory and/or inhibitory weights by an amount that goes from $\pm 1$ to $\pm 4$.
\end{itemize}
The score of each parameter set is defined as the Kendall-tau correlation between input and decoded parameters.

\subsection{Signal classification}
\label{sec:method_classification}
In Fig.~\ref{fig:backbone}, classification of signal type is performed using both time information and order information of the spiking encoding. We define as classification with time information a linear Support Vector Classifier (SVC) that considers the $N$-dimensional spiking encoding $\vec{y^\star}$, with $N$ the number of neurons. We define as classification with order information a linear SVC that considers a binary vector $\vec{o^\star}$ of dimension $N(N-1)/2$. Each entry in $\vec{o^\star}$ considers the relation between neuron $i$ and neuron $j$. An entry is equal to $1$ if neuron $i$ precedes neuron $j$ in $\vec{y^\star}$, and it is equal to $0$ otherwise. In Fig.~\ref{fig:classification}
only classification with order information is used. In Fig.~\ref{fig:classification}, the linear SVC accuracy obtained with order information is compared with the linear SVC accuracy obtained using the continuous non-encoded signal.

\subsection{Stereotyped spiking sequences}
\label{sec:method_backbone}
For each signal type and network size, we compute the mean spiking sequence over all stimuli. The neurons are then re-ordered based on their mean spike time in ascending order. 
The consistency of the spiking sequences is assessed with the Mutation Index (MI), defined as the mean Kendall-tau correlation between the mean spiking sequence and the single-trial sequences:

\begin{equation}
MI = \frac{1}{T} \sum_{t=1}^{T} \tau(S_{\text{mean}}, S_t)
\end{equation}

where:
\begin{itemize}
    \item $S_{\text{mean}}$ is the mean spiking sequence,
    \item $S_t$ is the spiking sequence of trial $t$,
    \item $\tau(S_{\text{mean}}, S_t)$ is the Kendall-tau correlation coefficient between the mean sequence and trial $t$,
    \item $T$ is the total number of trials.
\end{itemize}
A value of MI close to 1 indicates highly reproducible spike sequences, while lower values suggest greater spiking order variability across trials.\\
To test the independence of the spiking pattern on the network size, we first computed the spiking sequences for all network sizes and repetitions. We then created, for each network size, one histogram with the spike times of all sequences. After scaling, we removed the mean from each histogram and computed the cosine similarity between histograms.

\subsection{Robustness}
\label{sec_method_robustness}
The robustness of the encoding is tested by making the trained network progressively more homogeneous. Time constants homogeneity is increased by randomly selecting one neuron in the population and setting its time constant values as the mean of the time constants of the population. Weight homogeneity follows the same reasoning, setting the weight of randomly selected neurons as the mean of the weights of the population.

The robustness of the encoding is also tested against random temporal jitters in the spike times. After recording all spike times for all stimuli, we applied temporal jitters sampled from $\mathcal{N}(0,\sigma)$ with $\sigma = [0.2,0.4,0.8,1]\ \text{ms}$ to the training and/or test set. We then trained the linear decoder and assessed the performance with the Kendall-tau correlation. 

Robustness to spike deletion is assessed by removing spikes from randomly selected neurons for each stimulus encoding. After spike removal, the activity of the population is re-referenced to the new median of active neurons' spike times and a linear decoder is trained and tested on the perturbed spiking sequences.

Results presented in Fig.~\ref{fig:robustness} are obtained from experiments with DoubleGauss signals and 4 different hardware networks.

\subsection{Statistics}
All on-chip regression results were repeated on 6 different hardware networks in different days to ensure robustness. The on-chip classification is performed using one single chip. Simulation regression and classification results are obtained from 10 different runs for each network size and signal type. The error bars in the figures are computed as one standard deviation.

\ack
Part of this work has been done within the 2024 CapoCaccia Neuromorphic Workshow. C.D.L has received funding from Bridge Fellowship founded by the Digital Society Initiative at University of Zurich (grant no.G-95017-01-12); F.C. was funded by the Swiss National Science Foundation (grant no. 204651) and received UZH Candoc Grant (grant no. FK-24-025). We thank Prof. Dr. Giacomo Indiveri and Prof. Dr. Johannes Sarnthein for the valuable guidance, support and useful discussions.
\newpage
\section*{References}
\bibliographystyle{unsrt}
\bibliography{spiking_encoder}

\begin{thebibliography}{10}

\bibitem{hopfield1995pattern}
John~J Hopfield.
\newblock Pattern recognition computation using action potential timing for stimulus representation.
\newblock {\em Nature}, 376(6535):33--36, 1995.

\bibitem{chong2020manipulating}
Edmund Chong, Monica Moroni, Christopher Wilson, Shy Shoham, Stefano Panzeri, and Dmitry Rinberg.
\newblock Manipulating synthetic optogenetic odors reveals the coding logic of olfactory perception.
\newblock {\em Science}, 368(6497):eaba2357, 2020.

\bibitem{chelaru2008efficient}
Mircea~I Chelaru and Valentin Dragoi.
\newblock Efficient coding in heterogeneous neuronal populations.
\newblock {\em Proceedings of the National Academy of Sciences}, 105(42):16344--16349, 2008.

\bibitem{habashy2024adapting}
Karim~G Habashy, Benjamin~D Evans, Dan~FM Goodman, and Jeffrey~S Bowers.
\newblock Adapting to time: why nature evolved a diverse set of neurons.
\newblock {\em arXiv preprint arXiv:2404.14325}, 2024.

\bibitem{mainen1995reliability}
Zachary~F Mainen and Terrence~J Sejnowski.
\newblock Reliability of spike timing in neocortical neurons.
\newblock {\em Science}, 268(5216):1503--1506, 1995.

\bibitem{luczak2013gating}
Artur Luczak, Peter Bartho, and Kenneth~D Harris.
\newblock Gating of sensory input by spontaneous cortical activity.
\newblock {\em Journal of Neuroscience}, 33(4):1684--1695, 2013.

\bibitem{luczak2015packet}
Artur Luczak, Bruce~L McNaughton, and Kenneth~D Harris.
\newblock Packet-based communication in the cortex.
\newblock {\em Nature Reviews Neuroscience}, 16(12):745--755, 2015.

\bibitem{vaz2023backbone}
Alex~P Vaz, John~H Wittig~Jr, Sara~K Inati, and Kareem~A Zaghloul.
\newblock Backbone spiking sequence as a basis for preplay, replay, and default states in human cortex.
\newblock {\em Nature Communications}, 14(1):4723, 2023.

\bibitem{xie2024neuronal}
Weizhen Xie, John~H Wittig~Jr, Julio~I Chapeton, Mostafa El-Kalliny, Samantha~N Jackson, Sara~K Inati, and Kareem~A Zaghloul.
\newblock Neuronal sequences in population bursts encode information in human cortex.
\newblock {\em Nature}, pages 1--8, 2024.

\bibitem{perez2021neural}
Nicolas Perez-Nieves, Vincent~CH Leung, Pier~Luigi Dragotti, and Dan~FM Goodman.
\newblock Neural heterogeneity promotes robust learning.
\newblock {\em Nature communications}, 12(1):5791, 2021.

\bibitem{she2021sequence}
Xueyuan She, Saurabh Dash, and Saibal Mukhopadhyay.
\newblock Sequence approximation using feedforward spiking neural network for spatiotemporal learning: Theory and optimization methods.
\newblock In {\em International Conference on Learning Representations}, 2021.

\bibitem{gast2024neural}
Richard Gast, Sara~A Solla, and Ann Kennedy.
\newblock Neural heterogeneity controls computations in spiking neural networks.
\newblock {\em Proceedings of the National Academy of Sciences}, 121(3):e2311885121, 2024.

\bibitem{zeldenrust2021efficient}
Fleur Zeldenrust, Boris Gutkin, and Sophie Den{\'e}ve.
\newblock Efficient and robust coding in heterogeneous recurrent networks.
\newblock {\em PLoS computational biology}, 17(4):e1008673, 2021.

\bibitem{gerwinn2009bayesian}
Sebastian Gerwinn, Jakob~H Macke, and Matthias Bethge.
\newblock Bayesian population decoding of spiking neurons.
\newblock {\em Frontiers in computational neuroscience}, 3:648, 2009.

\bibitem{lazar2004time}
Aurel~A Lazar.
\newblock Time encoding with an integrate-and-fire neuron with a refractory period.
\newblock {\em Neurocomputing}, 58:53--58, 2004.

\bibitem{lazar2009reconstruction}
Aurel~A Lazar and Eftychios~A Pnevmatikakis.
\newblock Reconstruction of sensory stimuli encoded with integrate-and-fire neurons with random thresholds.
\newblock {\em EURASIP Journal on Advances in Signal Processing}, 2009:1--14, 2009.

\bibitem{adam2020sampling}
Karen Adam, Adam Scholefield, and Martin Vetterli.
\newblock Sampling and reconstruction of bandlimited signals with multi-channel time encoding.
\newblock {\em IEEE Transactions on Signal Processing}, 68:1105--1119, 2020.

\bibitem{schrauwen2003bsa}
Benjamin Schrauwen and Jan Van~Campenhout.
\newblock Bsa, a fast and accurate spike train encoding scheme.
\newblock In {\em Proceedings of the International Joint Conference on Neural Networks, 2003.}, volume~4, pages 2825--2830. IEEE, 2003.

\bibitem{rabinovich2001dynamical}
M~Rabinovich, A~Volkovskii, P~Lecanda, R~Huerta, HDI Abarbanel, and G~Laurent.
\newblock Dynamical encoding by networks of competing neuron groups: winnerless competition.
\newblock {\em Physical review letters}, 87(6):068102, 2001.

\bibitem{buonomano1995temporal}
Dean~V Buonomano and Michael~M Merzenich.
\newblock Temporal information transformed into a spatial code by a neural network with realistic properties.
\newblock {\em Science}, 267(5200):1028--1030, 1995.

\bibitem{stanojevic2024high}
Ana Stanojevic, Stanis{\l}aw Wo{\'z}niak, Guillaume Bellec, Giovanni Cherubini, Angeliki Pantazi, and Wulfram Gerstner.
\newblock High-performance deep spiking neural networks with 0.3 spikes per neuron.
\newblock {\em Nature Communications}, 15(1):6793, 2024.

\bibitem{goltz2021fast}
Julian G{\"o}ltz, Laura Kriener, Andreas Baumbach, Sebastian Billaudelle, Oliver Breitwieser, Benjamin Cramer, Dominik Dold, Akos~Ferenc Kungl, Walter Senn, Johannes Schemmel, et~al.
\newblock Fast and energy-efficient neuromorphic deep learning with first-spike times.
\newblock {\em Nature machine intelligence}, 3(9):823--835, 2021.

\bibitem{deneve2016efficient}
Sophie Den{\`e}ve and Christian~K Machens.
\newblock Efficient codes and balanced networks.
\newblock {\em Nature neuroscience}, 19(3):375--382, 2016.

\bibitem{boerlin2013predictive}
Martin Boerlin, Christian~K Machens, and Sophie Den{\`e}ve.
\newblock Predictive coding of dynamical variables in balanced spiking networks.
\newblock {\em PLoS computational biology}, 9(11):e1003258, 2013.

\bibitem{frady2019robust}
E~Paxon Frady and Friedrich~T Sommer.
\newblock Robust computation with rhythmic spike patterns.
\newblock {\em Proceedings of the National Academy of Sciences}, 116(36):18050--18059, 2019.

\bibitem{deckers2022extended}
Lucas Deckers, Ing~Jyh Tsang, Werner Van~Leekwijck, and Steven Latr{\'e}.
\newblock Extended liquid state machines for speech recognition.
\newblock {\em Frontiers in Neuroscience}, 16:1023470, 2022.

\bibitem{maass2002real}
Wolfgang Maass, Thomas Natschl{\"a}ger, and Henry Markram.
\newblock Real-time computing without stable states: A new framework for neural computation based on perturbations.
\newblock {\em Neural computation}, 14(11):2531--2560, 2002.

\bibitem{moradi2017scalable}
Saber Moradi, Ning Qiao, Fabio Stefanini, and Giacomo Indiveri.
\newblock A scalable multicore architecture with heterogeneous memory structures for dynamic neuromorphic asynchronous processors (dynaps).
\newblock {\em IEEE transactions on biomedical circuits and systems}, 12(1):106--122, 2017.

\bibitem{zendrikov2023brain}
Dmitrii Zendrikov, Sergio Solinas, and Giacomo Indiveri.
\newblock Brain-inspired methods for achieving robust computation in heterogeneous mixed-signal neuromorphic processing systems.
\newblock {\em Neuromorphic Computing and Engineering}, 3(3):034002, 2023.

\bibitem{chase2007first}
Steven~M Chase and Eric~D Young.
\newblock First-spike latency information in single neurons increases when referenced to population onset.
\newblock {\em Proceedings of the National Academy of Sciences}, 104(12):5175--5180, 2007.

\bibitem{costa2024robust}
Filippo Costa, Eline~V Schaft, Geertjan Huiskamp, Erik~J Aarnoutse, Maryse~A van’t Klooster, Niklaus Krayenb{\"u}hl, Georgia Ramantani, Maeike Zijlmans, Giacomo Indiveri, and Johannes Sarnthein.
\newblock Robust compression and detection of epileptiform patterns in ecog using a real-time spiking neural network hardware framework.
\newblock {\em Nature Communications}, 15(1):3255, 2024.

\bibitem{chicca2014neuromorphic}
Elisabetta Chicca, Fabio Stefanini, Chiara Bartolozzi, and Giacomo Indiveri.
\newblock Neuromorphic electronic circuits for building autonomous cognitive systems.
\newblock {\em Proceedings of the IEEE}, 102(9):1367--1388, 2014.

\end{thebibliography}

\end{document}